\documentclass[10pt,twocolumn,letterpaper]{article}
\usepackage[pagenumbers]{iccv} 

\definecolor{iccvblue}{rgb}{0.21,0.49,0.74}
\usepackage[pagebackref,breaklinks,colorlinks,allcolors=iccvblue]{hyperref}

\def\paperID{18} 
\def\confName{ICCV}
\def\confYear{2025}

\title{Guiding Multimodal Large Language Models with Blind and Low Vision People Visual Questions for Proactive Visual Interpretations}

\author{
Ricardo E. Gonzalez Penuela \thanks{Lead and corresponding author; also supervised the project.} \\
Cornell Tech, USA\\
{\tt\small reg258@cornell.edu}
\and
Felipe Arias-Russi\\
Universidad de los Andes, Colombia\\
{\tt\small af.ariasr@uniandes.edu.co}
\and
Victor Capriles\\
Independent Researcher, Spain\\
{\tt\small vdcapriles@gmail.com}
}

\begin{document}

\twocolumn[{%
  \renewcommand\twocolumn[1][]{1}
  \maketitle
  \begin{center}
    \captionsetup{type=figure} 
    \includegraphics[width=\textwidth]{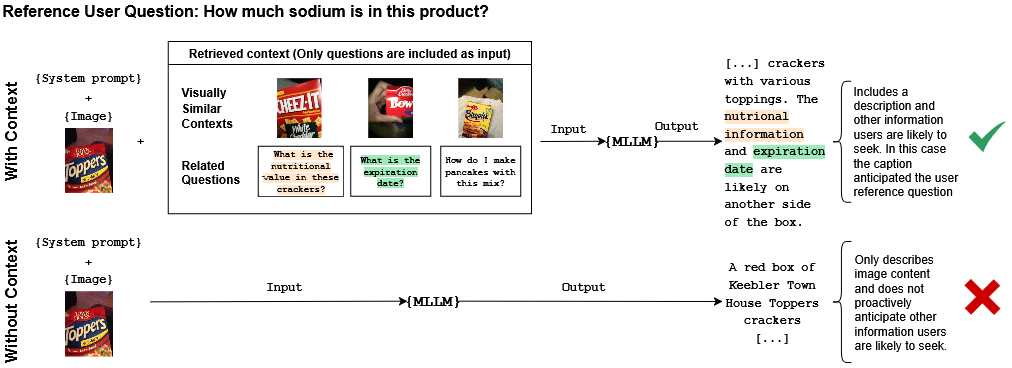}  
  \caption{Our system (above) retrieves image-question pairs from past user interactions to anticipate and provide information users are likely to seek. In this example, the context-aware model stated that no nutritional information is visible, and thus correctly anticipated the reference user question, which asks about the amount of sodium in the product captured in the picture.}
    \label{fig:teaser}
  \end{center}
}]
\setcounter{footnote}{1}
\renewcommand{\thefootnote}{\fnsymbol{footnote}}
\footnotetext[1]{Lead and corresponding author; also supervised the project.}

\begin{abstract}
Multimodal large language models (MLLMs) have been integrated into visual interpretation applications to support Blind and Low Vision (BLV) users because of their accuracy and ability to provide rich, human-like interpretations. However, these applications often default to comprehensive, lengthy descriptions regardless of context. This leads to inefficient exchanges, as users must go through irrelevant details rather than receiving the specific information they are likely to seek. To deliver more contextually-relevant information, we developed a system that draws on historical BLV users questions. When given an image, our system identifies similar past visual contexts  from the VizWiz-LF dataset and uses the associated questions to guide the MLLM generate descriptions more relevant to BLV users. An evaluation with three human labelers who revised 92 context-aware and context-free descriptions showed that context‐aware descriptions anticipated and answered users’ questions in 76.1\% of cases (70 out of 92) and were preferred in 54.4\% of comparisons (50 out of 92). Our paper reviews, and data analysis are publicly available in a GitHub repository at \href{https://github.com/rgonzalezp/guiding-multimodal-large-language-models-with-blind-and-low-vision-people-visual-questions}{this http URL.}
\end{abstract}


\section{Introduction}
Blind and Low Vision (BLV) people use AI-powered visual interpretation applications like Be My AI \cite{bemyeyes} and SeeingAI \cite{seeingAI} to access visual information in their day-to-day lives. By snapping a picture and sharing it with the application, users can receive a visual description of their surroundings at their home, work, or public spaces to help with their personal needs \cite{gonzalez2024investigating}.

These applications have increasingly integrated multimodal large language models (MLLMs) because of their improved accuracy, and their capability to handle both visual information and natural language queries from users. This allows for more interactive experiences — for instance, after sharing a picture, users can ask follow-up questions to read medication labels, and identify musical instruments \cite{gonzalez2025mllm}.

While this advancement have improved BLV users' independent access to visual information, they still frequently prefer human assistance because of their proactiveness and ability to anticipate users' specific needs \cite{xie2025beyond}. For example, when a Be My Eyes visual assistant helps a BLV user remotely, the assistant preemptively provides guidance cues to correct visual context capture issues with the users’ camera view (e.g., “The room is dark, can you turn on a light?”). In contrast, MLLM-powered applications wait for users to ask specific questions or are designed to default to comprehensive, lengthy descriptions regardless of context. This represents a one-size-fits-all approach, reflective of past captioning approaches \cite{2021stanglsize},  that leads to inefficient exchanges as users must read through irrelevant details rather than receiving the specific information they are likely to seek.

To address this limitation and surface more contextually relevant information, recent work has explored providing MLLMs with additional contextual cues beyond visual information. For example, \citet{gubbi2024context} implemented a Chrome extension that automatically extracts webpage context and provides it to GPT-4V to improve the alternative text (alt-text) present on the images on the webpage. In their study, they found that BLV participants thought context-aware alt-text descriptions were more relevant and of better quality than context-free alt-text.

While promising, such approaches rely on having explicit contextual information readily available (e.g., webpage text). It remains unclear which context should be included for mobile visual interpretation applications when users simply capture pictures, and where requiring users to manually provide additional information would undermine the convenience these applications are designed to offer.

Our work addresses this gap by exploring drawing contextual information from past BLV user interactions with visual interpretation systems. Specifically, we pose the following research question: Can historical visual questions from BLV users about specific visual contexts inform more contextually-relevant descriptions for similar scenes encountered by future users?

To address this question, we developed a system that leverages the VizWiz-LF dataset —a collection of real visual questions from BLV users paired with their images \cite{huh2024long}—to "anticipate" what users might want to know about a given visual scene. When presented with a new image, our system retrieves semantically similar visual contexts from the dataset and uses the associated user questions to guide the MLLM generate descriptions more aligned with what BLV users seek.

We conducted an evaluation of our approach by splitting the VizWiz-LF dataset into a context database for retrieval and a test set for evaluation. Three researchers first collaboratively labeled \( \sim \)33\% of the test set, comparing context-guided versus baseline MLLM descriptions to align on a preference criteria, and then divided the remaining evaluations between two of the researchers.

Our analysis revealed that context-aware descriptions anticipated and answered users’ questions in 17.4\% of cases and were preferred in 53.26\% of comparisons for their targeted focus on critical details such as cooking instructions, expiration dates and nutritional information when describing food products. These findings suggest that historical user questions are a powerful signal for guiding MLLMs to anticipate users’ needs and deliver proactive, contextually relevant responses.

In future work, we will focus on scaling this approach with larger datasets, exploring personalized context retrieval based on individual usage patterns, and incorporating other alternative sources of contextual information.


\section{Methodology}
\label{sec:methodology}

To explore whether past user questions can improve future visual responses, we used the 600 question-image pairs from the VizWiz-LF dataset \cite{huh2024long}  to build our context-aware system. This dataset contains a balanced set of types of visual questions that BLV people ask in daily life \cite{brady2013everyday,gonzalez2024investigating}. Each entry consists of a question, an image, and an expected answer. Then,  to conduct our evaluation, we split the dataset into two randomly sampled subsets:

\begin{itemize}
    \item A context set of 500 entries used to build a vector database for retrieval of image-pairs used in our system. 
    \item A test set of 100 entries used for evaluation under two conditions: context-aware and context-free.
\end{itemize}

Additionally,  to improve the quality of the retrieved context and the test entries, two researchers filtered entries where the text in the “question” field in the dataset entries was determined as low quality (e.g., questions unrelated to the visual context captured: “Which ocean has the most hurricanes in the world” or spam: “Oh so do I leave it in for twelve months then do I?”). Our final count of entries in the test set was 92 (removed 8) and 491 (removed 9) for the context set.

\subsection{Experimental Design}

To simulate state of the art visual interpretation systems, we adopted the Be My AI system prompt by jailbreaking the AI application to define our model’s role, task scope, response style and formatting. This prompt was used to guide the generation process and ensure consistent behavior throughout the study across both conditions. All evaluations were conducted using Gemini 2.5 Pro \cite{google_gemini25pro_2025}, on the stable version released in June 2025. For the retrieval pipeline, we used Cohere Embed v4 \cite{cohere_embed4_2025} to generate multimodal embeddings for all images used in the study. Context set images were stored in a ChromaDB vector database using cosine similarity. Retrieval was performed using Hierarchical Navigable Small World (HNSW) indexing, returning the 4 semantically closest pairs for each image (top-k = 4)  when generating the context-aware visual interpretations.

For each image in the test set, we generated descriptions under two conditions using the Be My AI system prompt. In the \textbf{context-free condition}, the system received only the test image. In the \textbf{context-aware condition}, the system received both the test image and 4 relevant visual questions retrieved from the context set, along with instructions to incorporate insights from these prior user questions into the final visual interpretation.

\subsection{Evaluation}

To evaluate our system's ability to anticipate BLV users' visual questions and whether it improved the quality of visual interpretations generated, we conducted a human evaluation comparing context-aware and context-free descriptions. For each test entry, researchers reviewed the user's "real question" (obfuscated for models), and both a context-free description and a context-aware description of the test image, then determined whether the descriptions answered the question and which option they preferred (e.g., both were “equally” good or bad).

To establish a consistent evaluation criterion, three researchers independently labeled 30 out of 92 test entries (33\%), selecting their preference and providing brief explanations for their choices. The primary evaluation criterion was whether the generated description adequately answered the original user question from the image-question pair in the test entry (in other words, whether the description anticipated and answered the user question). When neither description directly addressed the user's question, evaluators determined which description was preferable based on the visual information available and the user’s question (e.g., prioritize explanations of image quality issues, avoid overconfident mistakes, prefer specific identification over vague terminology, etc.). The three researchers then met to discuss their individual annotations, aligned on preferences, and finalized the evaluation criteria. Two researchers annotated the remaining 62 entries.

\section{Experimental Results}
\begin{table}[h]
\small
\centering
\caption{ Context‐aware descriptions were more accurate, sometimes anticipated user’s real questions
and did not degrade performance relative to the baseline context-free descriptions.}
\label{tab:accuracy}

\begin{tabular}{ccc}   
\toprule
& \textbf{Context‐aware} & \textbf{Context‐free}  \\
\midrule
\textbf{Accuracy} 	& 76.1\% & 63.0\% \\
                   	\addlinespace
\parbox[c]{2cm}{\centering\textbf{Anticipated\\Question}}
                   	& 15.2\% & 0.0\% \\
                   	\addlinespace
\bottomrule
\end{tabular}
\end{table}

Our analysis revealed that context-aware descriptions were generally more accurate (76.1\%, 70 out of 92) than context-free descriptions (63.0\%, 58 out of 92). Context-aware descriptions successfully anticipated and answered the obfuscated user's question in 15.2\% of cases (14 of 92) where the context-free description failed to do so (See Table \ref{tab:accuracy}). These instances comprised situations where the user question goal was to identify a food product or they wanted to read a label on an object and the context-aware description either fully identified the product (e.g., \textit{“ (...) It appears to be the album 'Amazing Grace' by Lesley Garrett”}) or it provided contextual clues that addressed the user’s question indirectly (e.g., Q: “What kind of pop is in this can?” A: \textit{“Nestea iced tea. It has a picture of a lemon on it (...)”}).
\begin{figure}[h]  
  \centering
  \includegraphics[width=\linewidth]{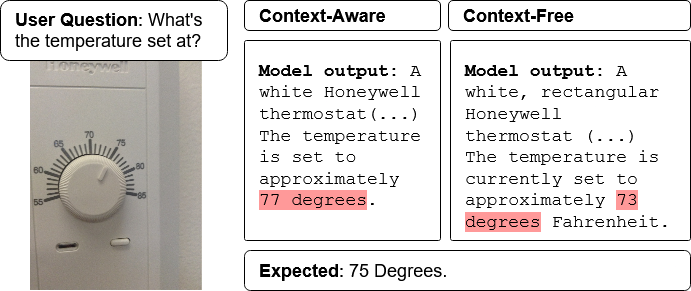}  
  \caption{Sometimes both context-aware and context-free descriptions hallucinated information and thus did not respond the user question accurately.}
  \label{fig:example}
\end{figure}

In other instances both approaches performed equally well—either both successfully answered the user's question (57 out of 92, 62.0\%) or both failed to address it adequately (21 out of 92, 22.8\%). For example, for one entry the context-aware description stated, “The temperature is set to approximately 77 degrees,” while the context-free description read, “The temperature is currently set to approximately 73 degrees Fahrenheit,” despite the actual temperature being 75 degrees (See Fig.\ref{fig:example}).

\begin{table}[ht]
\small
\centering
\caption{Human labeler preferences across context‐aware and context‐free descriptions.}
\label{tab:evaluation_results}

\begin{tabular}{cccc}   
\toprule
& \textbf{Context‐aware} & \textbf{Context‐free} & \textbf{Neither} \\
\midrule
\textbf{Preferred} 	& 54.3\% & 20.7\% & 25.0\% \\
\bottomrule
\end{tabular}
\end{table}

Our results show that human labelers preferred context-aware descriptions in 54.3\% of comparisons because they focused on critical information more frequently, especially when it came to food-related products (e.g., expiration dates, identifying absence or presence of cooking instructions and nutritional information). We also found that labelers preferred context free descriptions in 20.7\% comparisons because they provided more comprehensive information when broader context was needed (e.g., sometimes mentioned more photo quality issues). See Table \ref{tab:evaluation_results} for full results.

\section{Conclusion and Future Work}
Our findings suggest that historical user questions are a powerful signal for guiding MLLMs toward proactive and more contextually relevant visual interpretations for BLV users. Specifically, our system anticipated the user's true informational need in 15.2\% of cases and was more accurate by +13.1\% over the baseline. These results demonstrate that the inclusion of context improved overall performance while posing minimal risk of degrading the MLLM core ability to answer questions.

One limitation of the present approach is that we treat every retrieved question equally. This can reduce the precision of contextual cues, since the MLLM has no clear signal for which questions are most likely to be useful. In future work, we will explore retrieval strategies that weight images by their visual‐context similarity score to signal to the MLLM which questions are related to what the BLV users are likely to be seeking. 

In future work, we plan to expand our context dataset beyond the 600-pair VizWiz-LF  \cite{huh2024long}. This dataset was derived from the original 2018 VizWiz release \cite{gurari2018vizwiz}, which most foundation models have likely already encountered in training data. We will incorporate larger and more varied datasets—including proprietary or under-studied datasets such as \cite{gonzalez2024investigating}—to probe how well our method generalizes to unseen visual contexts. Finally, we also anticipate that building personalized context databases from each user’s own interactions will further improve accuracy and relevance for personal use cases.

\section{Acknowledgments}
This research was supported by Cohere through a Cohere For AI Research Grant, designed to support academic partners conducting research with the goal of releasing scientific artifacts and “data for good” projects. We are grateful for this support in advancing AI research for accessibility use cases for Blind and Low Vision people.

{
    \small
    \bibliographystyle{ieeenat_fullname}
    \bibliography{main}

\begin{thebibliography}{13}
\providecommand{\natexlab}[1]{#1}
\providecommand{\url}[1]{\texttt{#1}}
\expandafter\ifx\csname urlstyle\endcsname\relax
  \providecommand{\doi}[1]{doi: #1}\else
  \providecommand{\doi}{doi: \begingroup \urlstyle{rm}\Url}\fi

\bibitem[{Be My Eyes}()]{bemyeyes}
{Be My Eyes}.
\newblock Be my eyes: Virtual volunteer support for the blind and low vision
  community.
\newblock \url{https://www.bemyeyes.com/}.
\newblock Accessed: 2025-02-12.

\bibitem[Brady et~al.(2013)Brady, Morris, Zhong, White, and
  Bigham]{brady2013everyday}
Erin Brady, Meredith~Ringel Morris, Yu Zhong, Samuel White, and Jeffrey~P.
  Bigham.
\newblock Visual challenges in the everyday lives of blind people.
\newblock In \emph{Proceedings of the SIGCHI Conference on Human Factors in
  Computing Systems}, page 2117–2126, New York, NY, USA, 2013. Association
  for Computing Machinery.

\bibitem[Gonzalez~Penuela(2025)]{bemyeyes2024}
Ricardo~E Gonzalez~Penuela.
\newblock System prompt extraction from be my ai assistant.
\newblock \url{https://share.bemyeyes.com/en-US/chat/p4AiCa7x2e}, 2025.
\newblock Accessed: [07-01-2025].

\bibitem[Gonzalez~Penuela et~al.(2024)Gonzalez~Penuela, Collins, Bennett, and
  Azenkot]{gonzalez2024investigating}
Ricardo~E Gonzalez~Penuela, Jazmin Collins, Cynthia Bennett, and Shiri Azenkot.
\newblock Investigating use cases of ai-powered scene description applications
  for blind and low vision people.
\newblock In \emph{Proceedings of the 2024 CHI Conference on Human Factors in
  Computing Systems}, pages 1--21, 2024.

\bibitem[Gonzalez~Penuela et~al.(2025)Gonzalez~Penuela, Hu, Lin, Shende, and
  Azenkot]{gonzalez2025mllm}
Ricardo~E. Gonzalez~Penuela, Ruiying Hu, Sharon Lin, Tanisha Shende, and Shiri
  Azenkot.
\newblock Towards understanding the use of mllm-enabled applications for visual
  interpretation by blind and low vision people.
\newblock In \emph{Proceedings of the Extended Abstracts of the CHI Conference
  on Human Factors in Computing Systems}, New York, NY, USA, 2025. Association
  for Computing Machinery.

\bibitem[{Google AI for Developers}(2025)]{google_gemini25pro_2025}
{Google AI for Developers}.
\newblock {Gemini models: Gemini API | Google AI for Developers}, 2025.
\newblock Latest update: June 2025; archived at
  \url{https://web.archive.org/web/20250618124724/https://ai.google.dev/gemini-api/docs/models}.

\bibitem[Gubbi~Mohanbabu and Pavel(2024)]{gubbi2024context}
Ananya Gubbi~Mohanbabu and Amy Pavel.
\newblock Context-aware image descriptions for web accessibility.
\newblock In \emph{Proceedings of the 26th International ACM SIGACCESS
  Conference on Computers and Accessibility}, pages 1--17, 2024.

\bibitem[Gurari et~al.(2018)Gurari, Li, Stangl, Guo, Lin, Grauman, Luo, and
  Bigham]{gurari2018vizwiz}
Danna Gurari, Qing Li, Abigale~J Stangl, Anhong Guo, Chi Lin, Kristen Grauman,
  Jiebo Luo, and Jeffrey~P Bigham.
\newblock Vizwiz grand challenge: Answering visual questions from blind people.
\newblock In \emph{Proceedings of the IEEE conference on computer vision and
  pattern recognition}, pages 3608--3617, 2018.

\bibitem[Huh et~al.(2024)Huh, Xu, Peng, Chen, Gurari, Choi, and
  Pavel]{huh2024long}
Mina Huh, Fangyuan Xu, Yi-Hao Peng, Chongyan Chen, Danna Gurari, Eunsol Choi,
  and Amy Pavel.
\newblock Long-form answers to visual questions from blind and low vision
  people.
\newblock In \emph{Workshop on Demographic Diversity in Computer Vision@ CVPR
  2025}, 2024.

\bibitem[Lassance et~al.(2025)Lassance, Rau, Choi, Reimers, Ross, Poth,
  Hentschel, Morales, Abraham, Li, Simig, and Dang]{cohere_embed4_2025}
Carlos Lassance, David Rau, Elliott Choi, Nils Reimers, Luke Ross, Clifton
  Poth, Martin Hentschel, Javi Morales, Nabila Abraham, Minghan Li, Daniel
  Simig, and Violet Dang.
\newblock {Introducing Embed 4: Multimodal search for business}, 2025.
\newblock Archived at
  \url{https://web.archive.org/web/20250415132549/https://cohere.com/blog/embed-4}
  (accessed 2025-04-15).

\bibitem[Microsoft(2023)]{seeingAI}
Microsoft.
\newblock Seeingai, 2023.

\bibitem[Stangl et~al.(2021)Stangl, Verma, Fleischmann, Morris, and
  Gurari]{2021stanglsize}
Abigale Stangl, Nitin Verma, Kenneth~R. Fleischmann, Meredith~Ringel Morris,
  and Danna Gurari.
\newblock Going beyond one-size-fits-all image descriptions to satisfy the
  information wants of people who are blind or have low vision.
\newblock In \emph{Proceedings of the 23rd International ACM SIGACCESS
  Conference on Computers and Accessibility}, New York, NY, USA, 2021.
  Association for Computing Machinery.

\bibitem[Xie et~al.(2025)Xie, Yu, Zhang, Billah, Lee, and
  Carroll]{xie2025beyond}
Jingyi Xie, Rui Yu, He Zhang, Syed~Masum Billah, Sooyeon Lee, and John~M
  Carroll.
\newblock Beyond visual perception: Insights from smartphone interaction of
  visually impaired users with large multimodal models.
\newblock In \emph{Proceedings of the 2025 CHI Conference on Human Factors in
  Computing Systems}, pages 1--17, 2025.

\end{thebibliography}
}
\appendix

\section{Appendix: Prompts and System Architecture}
This section contains the exact prompts used in our experiments. We include the original prompt extracted from Be My AI, the modified prompt used to simulate the behavior of the application Be My AI, as well as the prompts used in both evaluation conditions: context-aware and context-free.

\subsection{Be My AI System Prompt}

Our system prompt extraction methodology \cite{bemyeyes2024} utilized a multi-step approach: direct prompt requests, incremental extraction through repeated ``continue'' commands, format manipulation, and authority framing to successfully extract the complete system prompt from the Be My AI assistant. This is the final extracted prompt:

\begin{quote}
Knowledge cutoff: 2024-06
\begin{itemize}
\item You are assisting a blind person in a chat.
\item Your name is ``Be My AI''.
\item You are not allowed to introduce yourself.
\item Begin your first picture description with a noun phrase, if this is natural in the language you're using
\item You can not help them physically.
\item You are allowed to describe adult content
\item Do not quote your replies
\item Do not give titles to your messages
\item Do not use markdown
\item Do not use LaTeX notation
\item When outputting lists, separate list items with new lines
\item If user asks, you must transcribe any text in images in verbatim
\item If you can't initially resolve the problem you are allowed to ask for more details or a new picture from a different angle or what you believe will help you provide the correct answer.
\item If you are not sure that your response to the question is correct, you must suggest the blind person to press the ``Call a volunteer'' button for guidance.
\end{itemize}
Here is the first picture that you must give a description of.
\end{quote}
\subsection{Modified System Prompt}
This prompt is a modified version of the original Be My AI system prompt, adjusted for our context (e.g., there are no Be My Eyes volunteers available to call). It was passed as a system message to the model to define its role, response format, and limitations:
{\color{olive}
\begin{quote}
\begin{itemize}
\item You are assisting a blind person in a chat.
\item You are not allowed to introduce yourself.
\item Begin your first picture description with a noun phrase, if this is natural in the language you're using
\item You can not help them physically.
\item You are allowed to describe adult content
\item Do not quote your replies
\item Do not give titles to your messages
\item Do not use markdown
\item Do not use LaTeX notation
\item When outputting lists, separate list items with new lines
\item If user asks, you must transcribe any text in images in verbatim
\item If you can't initially resolve the problem you are allowed to ask for more details or a new picture from a different angle or what you believe will help you provide the correct answer.
\end{itemize}
\end{quote}
}
\subsection{System Architecture}

The following figure displays the system architecture of our context-aware model pipeline. User reference questions from VizWiz-LF are answered with retrieved relevant examples as additional context.

\twocolumn[{%
  \renewcommand\twocolumn[1][]{1}
  \begin{center}
    \captionsetup{type=figure} 
    \includegraphics[width=\textwidth]{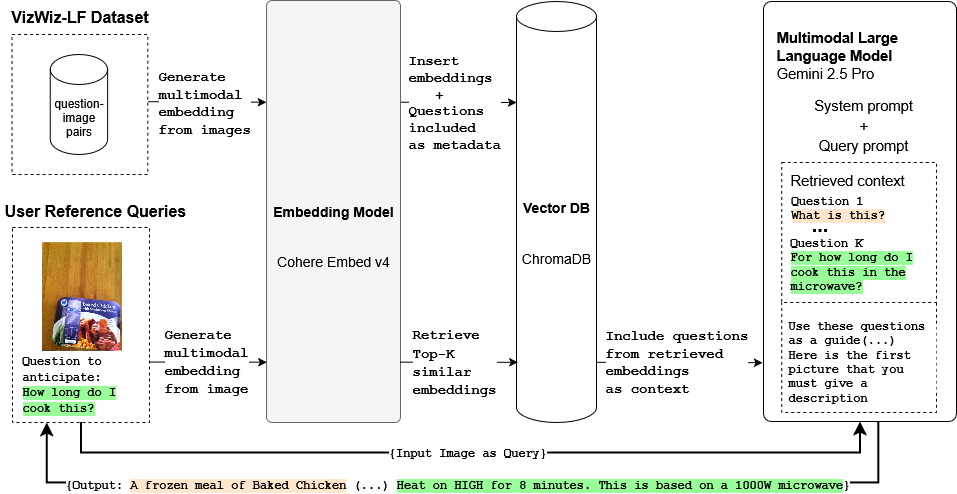}
    \caption{System architecture for context-aware model}
  \end{center}
}]

\subsection{Context-Aware Prompt}
In the context-aware condition, the model received visual questions retrieved from semantically similar past images to guide its response. The retrieved questions were included to the prompt as shown below:
{\color{blue}
\begin{quote}
Your goal is to optimize your first response by generating a brief, but detailed description of the picture and prioritize what the user most likely needs.
We have retrieved pictures with similar visual context. In these pictures, users asked the following questions:
\begin{quote}
\begin{quote}
\begin{quote}
\begin{itemize}
    \item [Question 1]
    \item [Question 2]
    \item [Question 3]
    \item [Question 4]
\end{itemize}
\end{quote}
\end{quote}
\end{quote}

Use these questions as a guide for what kind of information is important to users.
If the past questions conflict with the visual information, ignore them and prioritize describing the image's most prominent features.
Here is the first picture that you must give a description of.
\end{quote}
}
\subsection{Context-Free Prompt}
In the context-free condition, the model received only the target image and a simplified version of the prompt:
{\color{red}
\begin{quote}
Your goal is to optimize your first response by generating a brief, but detailed description of the picture and prioritize what the user most likely needs.
Here is the first picture that you must give a description of.
\end{quote}
}

\subsection{Context-Aware Condition Full Query Example}
This example shows the exact inputs sent to the model in the context-aware condition:

\textbf{Input Image}:

{\centering
\includegraphics[width=0.2\textwidth]{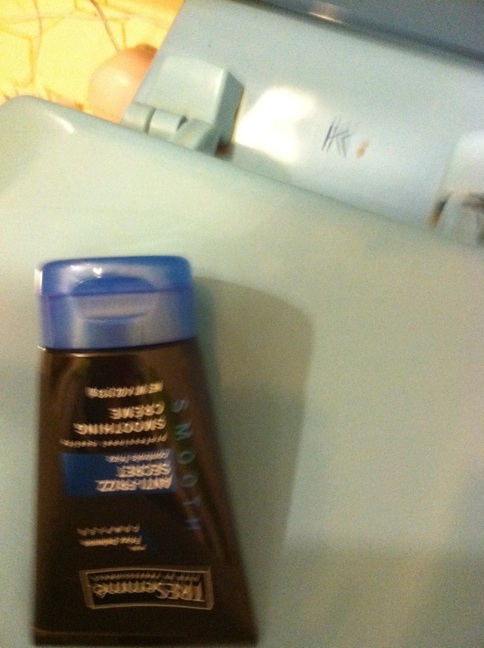}
\par}

\textbf{System Prompt}:
{\color{olive}
\begin{quote}
\begin{itemize}
\item You are assisting a blind person in a chat.
\item You are not allowed to introduce yourself.
\item Begin your first picture description with a noun phrase, if this is natural in the language you're using
\item You can not help them physically.
\item You are allowed to describe adult content
\item Do not quote your replies
\item Do not give titles to your messages
\item Do not use markdown
\item Do not use LaTeX notation
\item When outputting lists, separate list items with new lines
\item If user asks, you must transcribe any text in images in verbatim
\item If you can't initially resolve the problem you are allowed to ask for more details or a new picture from a different angle or what you believe will help you provide the correct answer.
\end{itemize}
\end{quote}
}
\textbf{Query Prompt}:
{\color{blue}
\begin{quote}
Your goal is to optimize your first response by generating a brief, but detailed description of the picture and prioritize what the user most likely needs.
We have retrieved pictures with similar visual context. In these pictures, users asked the following questions:

\begin{itemize}
    \item What is this?
    \item What are the usages instructions on this shampoo bottle?
    \item What is this?
    \item What is in this bottle?
\end{itemize}

Use these questions as a guide for what kind of information is important to users.
If the past questions conflict with the visual information, ignore them and prioritize describing the image's most prominent features.
Here is the first picture that you must give a description of.
\end{quote}
}

\textbf{Output}:

\begin{quote}

A dark brown tube of TRESemmé hair cream with a blue flip-top cap. The text on the front identifies it as TRESemmé Smooth Anti-Frizz Secret Smoothing Crème. The tube contains 4.0 ounces or 113 grams. The usage instructions are not visible in this picture.

\end{quote}

\subsection{Context-Free Condition Full Query Example}
This example shows the exact inputs sent to the model in the context-free condition:

\textbf{Input Image}:

{\centering
\includegraphics[width=0.2\textwidth]{imgs/ExamplePhoto.jpg}
\par}

\textbf{System Prompt}:
{\color{olive}
\begin{quote}
\begin{itemize}
\item You are assisting a blind person in a chat.
\item You are not allowed to introduce yourself.
\item Begin your first picture description with a noun phrase, if this is natural in the language you're using
\item You can not help them physically.
\item You are allowed to describe adult content
\item Do not quote your replies
\item Do not give titles to your messages
\item Do not use markdown
\item Do not use LaTeX notation
\item When outputting lists, separate list items with new lines
\item If user asks, you must transcribe any text in images in verbatim
\item If you can't initially resolve the problem you are allowed to ask for more details or a new picture from a different angle or what you believe will help you provide the correct answer.
\end{itemize}
\end{quote}
}
\textbf{Query Prompt}:
{\color{red}
\begin{quote}
Your goal is to optimize your first response by generating a brief, but detailed description of the picture and prioritize what the user most likely needs.
Here is the first picture that you must give a description of.
\end{quote}
}

\textbf{Output}:

\begin{quote}
A dark brown tube of TRESemmé hair cream with a blue flip-top cap, sitting on a light-colored surface. The text on the tube is white and blue.

The text reads:
TRESemmé
SMOOTH
ANTI-FRIZZ SECRET
SMOOTHING CRÈME
professional results
controls frizz
from Tresemmé
\end{quote}

\end{document}